\pdfoutput=1
\documentclass[10pt,twocolumn,letterpaper]{article}

\usepackage[pagenumbers]{cvpr} % To force page numbers, e.g. for an arXiv version

\definecolor{cvprblue}{rgb}{0.21,0.49,0.74}
\usepackage[pagebackref,breaklinks,colorlinks,allcolors=cvprblue]{hyperref}

\def\paperID{} % *** Enter the Paper ID here
\def\confName{CVPR}
\def\confYear{2025}
\usepackage{algorithmic}
\usepackage{algorithm}

\usepackage{makecell}
\usepackage{diagbox}  % Package for diagonal line in tables
\usepackage{array}
\usepackage{graphicx}

\title{FedDPC : Handling Data Heterogeneity and Partial Client Participation in Federated Learning}

\author{Mrinmay Sen \\
Indian Institute of Technology, Hyderabad\\
Hyderabad, India\\
{\tt\small senmrinmay@alumni.iith.ac.in}
\and
Subhrajit Nag\\
Telecom Sudparis\\
France\\
{\tt\small subrajit.nag@telecom-sudparis.eu}
}

\begin{document}
\maketitle
\begin{abstract}
Data heterogeneity is a significant challenge in modern federated learning (FL) as it creates variance in local model updates, causing the aggregated global model to shift away from the true global optimum. Partial client participation in FL further exacerbates this issue by skewing the aggregation of local models towards the data distribution of participating clients. This creates additional variance in the global model updates, causing the global model to converge away from the optima of the global objective. These variances lead to instability in FL training, which degrades global model performance and slows down FL training. While existing literature primarily focuses on addressing data heterogeneity, the impact of partial client participation has received less attention. In this paper, we propose FedDPC, a novel FL method, designed to improve FL training and global model performance by mitigating both data heterogeneity and partial client participation. FedDPC addresses these issues by projecting each local update onto the previous global update, thereby controlling variance in both local and global updates. To further accelerate FL training, FedDPC employs adaptive scaling for each local update before aggregation. Extensive experiments on image classification tasks with multiple heterogeneously partitioned datasets validate the effectiveness of FedDPC. The results demonstrate that FedDPC outperforms state-of-the-art FL algorithms by achieving faster reduction in training loss and improved test accuracy across communication rounds.
\end{abstract}    
\section{Introduction}
\label{sec:intro}

Federated learning (FL) constitutes a finite-sum optimization problem in which numerous data sources or clients collaboratively learn a shared global model by optimizing a global objective function, which is the average of all the clients' objectives. This collaborative nature of FL eliminates the need to centralize local data on a common server, thereby safeguarding the privacy of individual data sources and providing economic benefits when training machine learning or deep learning models on large volumes of data. FedAvg \cite {fedavgmcmahan2017} is the first FL algorithm, where in each communication round, each of the clients updates the shared global model with their local data and local stochastic gradient descent (SGD) optimizer \cite{sgdketkar2017} and the hosting server collects all the locally trained models and aggregates them to find the updated global model. This FL iteration is continued until some targeted performance is achieved from the global model. FedAvg shows promising performance, when data are homogeneously distributed across all the clients \cite{huang2021personalized}. When data are heterogeneous across clients, the performance of FedAvg gets declined.

\begin{figure}[!h]
  \centering
  \includegraphics[width=0.8\linewidth]{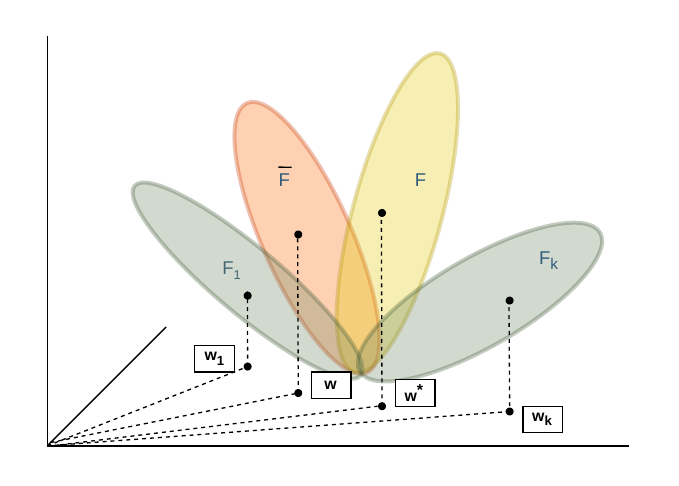}
  \caption{This figure shows the effect of client-drift in FL. Due to client-drift, the global model is optimized at the optima ($\textbf{w}$) of a different objective $\Bar{F}$ instead of the optima ($\textbf{w}^*$) of the global objective $F$ and each local model $\textbf{w}_i$ is optimized away from the global optima as well as optima of others client's objective $F_j$, where $i \neq j$.}
  \label{CD}
\end{figure}

Data heterogeneity across clients creates variance in local model updates, which causes each locally trained model to drift from the optima of the other clients' local objectives as well as the global objective, a phenomenon known as client-drift \cite{karimireddy2020scaffold}. The aggregation of these diverse local updates results in objective inconsistency in federated learning, causing the aggregated model, or the global model, to be optimized at a point that is not the true optimum of the global objective \cite{wang2020tackling}. This inconsistency of local and global updates is shown in Figure \ref{CD}. Due to data heterogeneity, each local model gets optimized at a point ($\textbf{w}_i$) of the local objective $F_i$, deviating from other local models ($\textbf{w}_j$, where $i\neq j$) and also from the optimal point ($\textbf{w}^*$) of the global objective $F$. This results in optimization of the global model $\textbf{w}$ at the optima of a different objective $\Bar{F}$, and not the true global objective. This issue in FL results in slow and unstable learning, as well as performance degradation of the global model \cite{kairouz2021issueoffl,li2019convergence,tan2022towardspfl,reddi2020adaptfedopt}. Coupled with the data heterogeneity issue in FL, another real-world problem arises from partial client participation due to device-level heterogeneity (e.g., unreliable network connections, power sources, etc.), limiting the availability of all clients in each FL communication round \cite{karimireddy2020scaffold,kairouz2021issueoffl}. Partial participation of heterogeneous local clients leads to different sets of local clients participating in each communication round, thereby skewing the aggregation of local updates toward the data distribution of the current set of participating clients. This bias in aggregation creates additional variance in the global model updates, causing the global model to converge away from the optimum of the global objective, which represents the combined data distribution of all clients. Coupled with the variance in local updates introduced by data heterogeneity, this additional variance in the global model further affects the training of federated learning \cite{kairouz2021issueoffl}. While existing literature primarily focuses on addressing data heterogeneity, the impact of partial client participation has received less attention. Although some works have been proposed to mitigate both issues in FL, there is still scope for improvement in FL training and global model performance, under heterogeneous and partial client participation settings. 

In this paper, we propose FedDPC to improve FL training and global model performance by mitigating the issues of heterogeneity and partial client participation in FL. FedDPC addresses these challenges by projecting each local update orthogonally onto the previous global update and modifying the local update with the residual update, the difference between the original local update and the projected update. Modifying the local updates with this residual ensures that the changes in the local updates are orthogonal to the previous global update, thereby ensuring an optimal change \cite{farajtabar2020orthogonal} in the local updates with respect to the global update of the previous communication round. This approach also helps to control the variance of the global update, as the new global update, the average of the local updates, remains orthogonal to the previous global update. 

To further accelerate FL training, FedDPC incorporates an adaptive scaling for each modified local update (i.e., the residual). This scaling considers the relative magnitude of the original local update in relation to the residual. Using two sided learning rates or scaling factors for clients and the server has been shown to improve FL training \cite{yang2021achieving}. However, existing methods in this area typically employ a fixed scaling factor for all clients before contributing to the global update in the server, which may not effectively control the variance across client updates. This, in turn, may fail to control the variance of the global update formed after aggregating the local updates. In contrast, the adaptive scaling mechanism in FedDPC adjusts the local update based on its relevance to the global update, ensuring that the aggregation of these updates results in optimal changes compared to the previous global update. This leads to a reduction in the variance of global updates across communication rounds, facilitating more stable and efficient global model training. The performance of FedDPC is validated through extensive experiments on image classification tasks using multiple heterogeneously partitioned datasets. The experimental results demonstrate that FedDPC outperforms state-of-the-art FL algorithms by accelerating FL training, achieving a faster reduction in training loss and improved test accuracy across communication rounds.

The primary contributions of this paper are as follows: 

\begin{itemize} 
    \item This paper proposes FedDPC, designed to improve FL training and global model performance by mitigating the issues of data heterogeneity and partial client participation in FL, which create variances in both local and global updates. \item To reduce the variance of local and global updates, FedDPC incorporates a projection-based modification of the local updates 
    \item To further accelerate FL training, FedHCF incorporates an adaptive scaling mechanism for the modified local updates. 
\end{itemize}

\section{Preliminaries}
In this section, we first show the problem formulation of FL. Then, we describe about Orthogonal Projection of a vector onto another vector.

\subsection{Problem Formulation}
Standard federated learning framework involves with finding optimized global model parameters $\textbf{w} \in \mathbb{R}^{d}$ by minimizing the global objective function $F(\textbf{w} ; \mathbb{D})$ as given in eq. \ref{eq:1}.
\begin{equation}
    \label{eq:1}
    \min_W F(\textbf{w} ; \mathbb{D}) = \frac{1}{k}\sum^k_{i=1} F_i(\textbf{w} ; \mathbb{D}_i).
\end{equation}
Where, $k$ denotes the number of participating clients, $F_i(\textbf{w} ; \mathbb{D}_i) =\frac{1}{|\mathbb{D}_i|}\sum^{|\mathbb{D}_i|}_{j=1} f_j(\textbf{w} ; \psi_j \in \mathbb{D}_i )$  is the local objective of the $i$-th client, considered as the local empirical loss, $f_j(.,.)$ is the local loss for the $j$-th sample, $\mathbb{D}_i$ is the dataset owned by the $i$-th client, and $\mathbb{D} = \{\mathbb{D}_1 \cup \mathbb{D}_2 \cup...\cup \mathbb{D}_k\}$.

\subsection{Orthogonal Projection}
To compute the projection of a vector \textbf{AB} onto another vector \textbf{AC}, a perpendicular line is drawn from \textbf{AB} to \textbf{AC}. Let \textbf{D} be the point where this perpendicular intersects \textbf{AC}. The segment \textbf{AD} is then defined as the orthogonal projection of \textbf{AB} onto \textbf{AC}, as expressed in Eq. \ref{eq:4}. Here, $(\cdot)$ denotes the dot product between two vectors.

\begin{equation}
    \label{eq:4}
    \textbf{AD}=\textbf{Proj}_{\textbf{AC}}({\textbf{AB}}) = \left(\frac{\textbf{AB} \cdot \textbf{AC}}{\textbf{AC} \cdot \textbf{AC}}\right) \textbf{AC}
\end{equation}

\section{Related Work}
\label{related_works}
Related works on handing data heterogeneity and partial clients participation in FL can be divided into five categories.

\textbf{Algorithms that use regularization term with local objective} include FedProx \cite{li2020fedprox}, FedDyn \cite{AcarZNMWS21fedydn}, FedSpeed \cite{Sun0H0T23fedspeed}, FedAlign \cite{mendieta2022localfedalign}, MOON \cite{li2021moon} etc. In FedProx, the local objective is restricted by adding a proximal term, thereby limiting the variability of the clients' updates. While optimizing the local objective, FedDyn incorporates a dynamically updated penalty term with the local sub-problem to align the global and local models within a certain limit. FedSpeed employs a prox-term to penalize the local offset and introduces a prox correction term to overcome the issue of incomplete local training caused by the use of this prox-term. The application of the prox-term and subsequent prox correction term results in additional computational load for the local clients. FedAlign applies a distillation-based regularization term to the local objective to limit the changes in local solutions. MOON incorporates contrastive learning during local training to reduce the distance between the locally learned models and the globally learned model.

\textbf{Algorithms that use two different scaling factors in the clients $\&$ server} include FedExP \cite{Jhunjhunwala0J23fedexp}, FedCM \cite{xu2021fedcm}, FedAMD \cite{Anchor_samplWu0QHLG23} etc. FedExP adaptively employs the server learning rate with the help of the extrapolation mechanism from the Projection Onto Convex Sets (POCS) algorithm. To mitigate the problem of partial participation of heterogeneous clients, FedCM leverages the previous step's global gradient to modify client gradient descent by incorporating a momentum-like term. FedAMD adopts anchor sampling to divide clients into anchor and miner groups. Anchor clients compute gradients with large batches, while miner clients refine updates with small batches.

\textbf{Algorithms that use previous step's local states} include SCAFFOLD \cite{karimireddy2020scaffold}, FedGA \cite{DandiBJ22fedga}, FedDC \cite{GaoFLC0022feddc}, MIFA \cite{gu2021fast}, FedVARP \cite{jhunjhunwala2022fedvarp} etc. SCAFFOLD incorporates variance reduction to address the problem of drift between local and global updates. FedGA relies on the displacement of the local gradient with respect to the global gradient when initializing local models. To limit the change of local models with respect to the global model, FedDC employs an auxiliary local drift variable. MIFA maintains a memory of the latest updates from all devices. In each communication round, it replaces the updates of the participating clients with the new ones and computes an average over all the updates stored in the memory. This approach helps account for the updates of inactive clients in each communication round. To overcome the issue of partial client participation, FedVARP adopts surrogate updates for all inactive clients in each communication round. These surrogate updates are formed by memorizing each client's most recent update.

\textbf{Algorithms that require some proxy data in the server} include the paper of Zhao et al. \cite{zhao2018emdmeasure}, FedBR \cite{guo2023fedbr} etc. These papers leverage the globally shared pseudo-data to address the local learning bias.

\textbf{Algorithms that involve with data generation in local clients using global generative model} include the work of Nagaraju et al. \cite{NagarajuSM23HandlingData}, FAug \cite{jeiong2018faug} etc. The work of Nagaraju et al. utilizes globally trained Gaussian mixture models to generate data in local clients, thereby helping to reduce the degree of heterogeneity across the clients. FAug trains a generative adversarial network on some samples of the minority classes, which are then shared with the server and uses this generative model to generate data in each local client.

Although the methods mentioned above outperform FedAvg in heterogeneous FL settings, each of them has its own limitations. These methods tend to be stateful, require more FL iterations, or impose privacy constraints. The use of regularization in local training, for example, can limit local convergence, resulting in the loss of novel information that could have otherwise contributed to the global model. Additionally, maintaining local states with low-rate partial client participation in FL presents significant challenges. Generating a global model and producing data on local clients also increases computational and memory overhead on the clients. Furthermore, making proxy data available on the server can be difficult and may raise privacy concerns for local clients. While existing literature primarily focuses on addressing data heterogeneity, the impact of partial client participation has received less attention. Although some works have been proposed to mitigate both issues in FL, there is still scope for improvement in FL training and global model performance under heterogeneous and partial client participation settings. 

In line with the second category of method i.e. \textbf{Algorithms that use two different scaling factors on the clients and server}, our proposed method, FedDPC, also employs two-sided scaling factors. However, our approach uniquely applies adaptive scaling to the projection-based variance-reduced local updates on the server. 

\section{Proposed Algorithm}
To overcome the effect of data heterogeneity and partial client participation in FL, we propose FedDPC (Handling \textbf{D}ata Heterogeneity and \textbf{P}artial \textbf{C}lient Participation in \textbf{Fed}erated learning). The outlined algorithm for the proposed FedDPC is presented in Algorithm. \ref{alg:algo1}.

\begin{algorithm}[!h]
   \caption{FedDPC}
   \label{alg:algo1}
\begin{algorithmic}[1]
    \item \textbf{Input:} $T$: FL communication rounds, $\textbf{w}_0$: Initial global model, $\eta_l$: local learning rate, $\eta_g$: Server learning rate, $\Delta_0\rightarrow 0$: Initial server update, $\lambda$ : Regularization term for adaptive scaling  \newline
    \FOR{$t=1$ {\bfseries to} $T$}
        \STATE Randomly pick a subset of clients $\overline{C} \subseteq C$
        \STATE Server broadcasts global model $\textbf{w}_{t-1}$ to all the available clients $\overline{C}$
        \STATE \underline{\textbf{In clients:}}\\
        \FOR{client j $\in \overline{C}$ \textbf{in parallel}}
            \STATE $\textbf{w}^0_{jt} \leftarrow \textbf{w}_{t-1}$
            \FOR{$m=1$ {\bfseries to} local iterations}
                \STATE Compute stochastic gradient $\textbf{g}^m_{jt}= \frac{\partial F_j(\textbf{w}^{m-1}_{jt}; \mathbb{D}^m_j)}{\partial \textbf{w}^{m-1}_{jt}}$ (Here, $\mathbb{D}^m_j \subseteq \mathbb{D}_j$)
                \STATE Update model $\textbf{w}^{m}_{jt} = \textbf{w}^{m-1}_{jt} - \eta_l \textbf{g}^m_{jt}$
            \ENDFOR
            \STATE $\textbf{w}_{jt} \leftarrow \textbf{w}^{m}_{jt}$
            \STATE Compute gradient update  $\boldsymbol{\Delta}_{jt} =\frac{\textbf{w}_{t-1} - \textbf{w}_{jt}}{\eta_l}$
            \STATE Broadcast $\boldsymbol{\Delta}_{jt}$ to the server
        \ENDFOR
        \STATE \underline{\textbf{In server:}}\\
        \STATE Receive set of available local updates $S=\{\boldsymbol{\Delta}_{jt}\}$
        \FOR{$\boldsymbol{\Delta}_{jt} \in S$}
            \item{ $\overline{\boldsymbol{\Delta}}_{jt}$= $\boldsymbol{\Delta}_{jt} - \textbf{Proj}_{\boldsymbol{\Delta}_{t-1}} {(\boldsymbol{\Delta}_{jt})}$ }
            \item{$\overline{\boldsymbol{\Delta}}_{jt} \leftarrow \left(\lambda + \frac{\|\boldsymbol{\Delta}_{jt}\|}{\|\overline{\boldsymbol{\Delta}}_{jt}\|}\right)\overline{\boldsymbol{\Delta}}_{jt}$}
        \ENDFOR

        \item {$\boldsymbol{\Delta}_t$ = $\frac{1} {|\overline{C}|} \sum\limits_{j \in \overline{C}} \overline{\boldsymbol{\Delta}}_{jt}$}

        \STATE Update the global model $\textbf{w}_{t}$ = $\textbf{w}_{t-1}$ - $\eta_g \boldsymbol{\Delta}_t$
    \ENDFOR
\end{algorithmic}
\end{algorithm}
%%%%%%%%%%%%%%%%%%%%%%%%%%%%%%%%%%%%%%%%%%%%%%%%%%%%%%%%%%%%%%%%%%%
\begin{table*}[ht]
\centering
\small % Use a larger font size
\setlength{\tabcolsep}{6pt} % Further reduce column spacing
\renewcommand{\arraystretch}{1.5} % Adjust row height for better readability
\begin{tabular}{|>{\centering\arraybackslash}p{1.2cm}|>{\centering\arraybackslash}p{1.2cm}|>{\centering\arraybackslash}p{1.7cm}|>{\centering\arraybackslash}p{1.0cm}|>{\centering\arraybackslash}p{1.7cm}|>{\centering\arraybackslash}p{1.7cm}|>{\centering\arraybackslash}p{1.4cm}|>{\centering\arraybackslash}p{1.7cm}|>{\centering\arraybackslash}p{1.4cm}|}
\hline
\textbf{Methods} & \multicolumn{5}{c|}{\textbf{Local costs}} & \multicolumn{3}{c|}{\textbf{Server costs}} \\
\cline{2-9}
& \textbf{Model initialization} & \textbf{Gradient computation} & \textbf{Model update} & \textbf{Computing gradient update} & \textbf{Total local costs} & \textbf{Global model update} & \textbf{Maintain previous state} & \textbf{Total server cost} \\
\hline
\textbf{FedAvg} &$O(d)$&$O(mnd)$&$O(md)$&-&$O(d+mnd+md)$&$O(k'd)$&$-$&$O(k'd)$\\
\hline
\textbf{FedProx} &$O(d)$&$O(mnd + md)$&$O(md)$&-&$O(d+mnd+2md)$&$O(k'd)$&$-$&$O(k'd)$\\
\hline
\textbf{FedExP} &$O(d)$&$O(mnd)$&$O(md)$&-&$O(d+mnd+md)$&$O(2k'd + 2d)$&$-$&$O(2k'd + 2d)$\\
\hline
\textbf{FedGA} &$O(3d)$&$O(mnd)$&$O(md)$&$O(d)$&$O(4d+mnd+md)$&$O(k'd)$&$O(k'd)$&$O(2k'd)$\\
\hline
\textbf{FedCM} &$O(d)$&$O(mnd)$&$O(2md)$&$O(d)$&$O(2d+mnd+2md)$&$O(k'd + d)$&$-$&$O(k'd+d)$\\
\hline
\textbf{FedVARP} &$O(d)$&$O(mnd)$&$O(md)$&$O(d)$&$O(2d+mnd+md)$&$O(2k'd + 2d)$&$O(d)$&$O(2k'd + 3d)$\\
\hline
\textbf{FedDPC} &$O(d)$&$O(mnd)$&$O(md)$&$O(d)$&$O(2d+mnd+md)$&$O(4k'd +d)$&$-$&$O(4k'd + d)$\\
\hline
\end{tabular}
\caption{Computation costs of various methods in both the server and the client, here, $n$ represents the number of local samples in each of the $m$ mini batchs and $k'$ is the number of participating clients in each round.}
\label{tab:Complex}
\end{table*}

%%%%%%%%%%%%%%%%%%%%%%%%%%%%%%%%%%%%%%%%%%%%%%%%%%%%%%%%%%%%%%%%
In FedDPC, server sends initial global model $\textbf{w}_{t-1}$ to all the available $\overline{C} \subseteq C$ clients (C is the set of all $k$ clients and $\overline{C}$ is the subset of clients,  which are randomly selected with equal probability for all the clients). Each client $j \in \overline{C}$ updates the shared global model by performing multiple stochastic gradient descent (SGD) steps on local data. After finding the updated local model $\textbf{w}_{jt}$, client $j$ broadcasts the model's update $\boldsymbol{\Delta}_{jt} \in \mathbb{R}^d = \frac{\textbf{w}_{t-1} - \textbf{w}_{jt}}{\eta_l}$ to the server, where $\eta_l$ is the local learning rate. Data heterogeneity creates variance in these local updates, causing the updates of local clients to diverge from each other and also from the global update. FedDPC addresses these challenges by projecting each local update orthogonally onto the previous global update and modifying the local update with the residual update, the difference between the original local update and the projected update. To further accelerate FL training, FedDPC incorporates a regularized adaptive scaling for each modified local update (i.e., the residual). After modification, FedDPC aggregates all the local updates and update the global model, which is then again sent to the next set of participating clients for retraining. 
\begin{figure}[!h]
  \centering
  \includegraphics[width=0.9\linewidth]{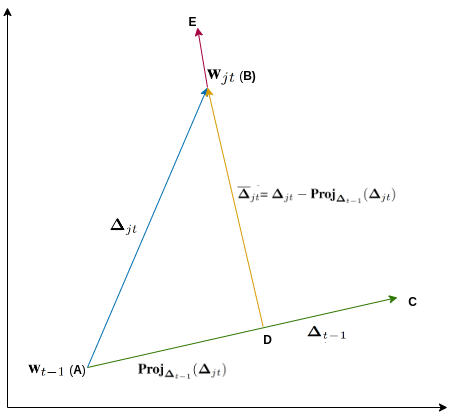}
  \caption{This figure describes the orthogonal projection of a local update onto the previous FL iteration's global update and use of this projection to correct this local update.}\label{f4}
\end{figure}

\subsection{Modify Local Update Using Orthogonal Projection}
Fig. \ref{f4} describes the modification of the local update of  $j$ - th client at $t$ - th FL communication round. Let, 
the point \textbf{A} denotes the global model $\textbf{w}_{t-1}$ at $(t-1)$ - th communication round, the point \textbf{B} denotes the updated local model $\textbf{w}_{jt}$, \textbf{AC} represents the global update $\boldsymbol{\Delta}_{t-1}$ at communication round $(t-1)$ and \textbf{AB} represents the local update $\boldsymbol{\Delta}_{jt}$. To reduce the variance of the local updates, FedDPC calculates the orthogonal projection of local update \textbf{AB} onto the previous iteration's global update \textbf{AC}, denoted by $\textbf{AD} = \textbf{Proj}_{\boldsymbol{\Delta}_{t-1}} (\boldsymbol{\Delta}_{jt})$ and then uses the residual \textbf{DB} =  $\boldsymbol{\overline{\Delta}}_{jt}$ = \textbf{AB} - \textbf{AD} = $\boldsymbol{\Delta}_{jt} - \textbf{Proj}_{\boldsymbol{\Delta}_{t-1}} (\boldsymbol{\Delta}_{jt})$ as the modification of the local update \textbf{AB}. Modifying the local updates with this residual ensures that the changes in the local updates are orthogonal to the previous global update \textbf{AC}, thereby ensuring an optimal change \cite{farajtabar2020orthogonal} in the local update with respect to the global update from the previous communication round, thus facilitating the reduction of variance in local updates. This approach also helps to control the variance of the global update, as the new global update, the average of the local updates, remains orthogonal to the previous global update.

\subsection{Adaptive Scaling of Local Update}

After modifying the local updates using orthogonal projection, FedDPC incorporates an adaptive scaling for each modified local update (i.e., the residual). This scaling considers the relative magnitude of the original local update in relation to the residual.  The adaptive scaling mechanism in FedDPC adjusts the local update based on its relevance to the global update, ensuring that the aggregation of these updates results in optimal changes compared to the previous global update. To achieve the same, FedDPC re-scales each of these modifications  adaptively from  DB to DE (Figure \ref{f4}) by using Eq. \ref{eq:5}, where $\lambda$ is a hyperparameter, which controls the degree of adaptiveness. The sensitivity of this $\lambda$ is analyzed in section \ref{Sen}.
\begin{equation}
    \label{eq:5}
    \boldsymbol{\overline{\Delta}}_{jt} \leftarrow \left(\lambda + \frac{\|\boldsymbol{\Delta_{jt}}\|}{\|\boldsymbol{\overline{\Delta}}_{jt}\|}\right)  \boldsymbol{\overline{\Delta}}_{jt}
\end{equation}
In Figure \ref{f4}, as $\textbf{DB} \perp \textbf{AD}$, $\angle \textbf{ADB} = 90^\circ$, indicating that the angle between the original local update \textbf{AB} and the previous global update \textbf{AC}, i.e., $\angle \textbf{BAD}$, is less than $90^\circ$. This implies that $ \text{cosec}\angle \textbf{BAD} = \frac{|\boldsymbol{\Delta_{jt}}|}{|\boldsymbol{\overline{\Delta}}_{jt}|}$, mapping from ($0^\circ$, $90^\circ$) to ($\infty$, 1). Thus, the adaptive scaling gives more weight to the update that is more relevant to the previous global update (relevance in terms of the angle between them), thereby helping to further control the changes of the local updates with respect to the global updates. Aggregation of these local updates ensures that the current global update does not deviate much from the previous global update, facilitating more stable and efficient global model training.
%%%%%%%%%%%%%%%%%%%%%%%%%%%%%%%%%%
%%%%%%%%%%%%%%%%%%%%%%%%%%%%%%%%%%
\subsection{Update the Global Model}

After performing adaptive scaling of all the modified local updates, FedDPC aggregates these $\{\boldsymbol{\overline{\Delta}}_{jt}\}$ and finds the global update $\boldsymbol{\Delta}_t$ as shown in Eq. \ref{eq:6} and uses this to update the global model $\textbf{w}_t$ as shown in Eq. \ref{eq:7}, where $\eta_g$ is the learning rate used in the server. 

\begin{equation}
    \label{eq:6}
   \boldsymbol{\Delta}_t = \frac{1} {|\overline{C}|} \sum\limits_{j \in \overline{C}} \boldsymbol{\overline{\Delta}}_{jt}
\end{equation}

\begin{equation}
    \label{eq:7}
    \textbf{w}_{t} = \textbf{w}_{t-1} - \eta_g \boldsymbol{\Delta}_t
\end{equation}

\subsection{Computation Cost}
The computation costs of our proposed FedDPC are analyzed in Table \ref{tab:Complex}. From this table, it can be noticed that the local computation cost is only $O(d)$ more than that of FedAvg and similar to FedCM and FedVARP. One issue with FedDPC is the increased server computation costs due to projection and adaptive scaling, which result in an increase of $O(3k' +d)$ computation costs compared to FedAvg, here $k'$ represents the number of participating clients in each round. But this may not be a big challenge if there are fewer participating clients in each round, as the server computation costs linearly depend on the number of participating clients $k'$. For a higher number of participating clients, FedDPC expects to have better computation resources from the server.

\section{Experiments}
This section describes the datasets used, the experimental setup, and the results for validating and comparing the performance of our proposed method, along with an analysis of the effects of hyperparameter and components. 

\subsection{Datasets}
We investigate the performance of FedDPC on three image classification datasets: CIFAR-10, CIFAR-100 \cite{krizhevsky2009learning}, and Tiny ImageNet \cite{chrabaszcz2017downsampled}. Both CIFAR-10 and CIFAR-100 consist of images of size $(3 \times 32 \times 32)$, with 50,000 training and 10,000 test samples. CIFAR-10 contains samples from 10 classes, while CIFAR-100 contains samples from 100 classes. Tiny ImageNet is a dataset of RGB $(3 \times 64 \times 64)$ images, consisting of 100,000 training and 10,000 test samples from 200 classes

We heterogeneously divide train data of each dataset into $k=100$ numbers of clients. To create heterogeneous partitions across clients, we use Dirichlet distribution, which is same as the implementation of the paper of Yurochkin et al. \cite{yurochkin2019bayesian}. We first sample $P_r \sim Dir_{k=100}(\alpha)$ and create $j^{th}$ client's dataset by allocating a $P_{(r,j)}$ proportion of the samples of $r^{th}$ class. For each dataset, we do experiments with two values of Dirichlet parameter $\alpha \in \{0.2, 0.6\}$, indicating two different labels of data heterogeneity across 100 clients.

\subsection{Experimental Setup}
This section gives an overview of the model used, the compared methods, performance metrics, and implementation details.

\subsubsection{Model}
For federated image classification of  CIFAR10 data, we use LeNet5 model \cite{lecun2015lenet} and for CIFAR100 and Tiny ImageNet, we use Resnet18 model \cite{he2016deep} with group normalization \cite{wu2018group}. 
\subsubsection{Performance Metrics}
To evaluate and compare the performance of our proposed method, we record the average training loss across clients and test accuracy( $= \frac{\text{correct predictions}}{\text{number of  samples}}$) of the global global in each communication round, along with the best test accuracy of the global model achieved across all communication rounds. The training loss in each communication round is measured by averaging the training loss across all participating clients. The test accuracy of the global model is calculated on the test data in each round. As a loss function, we use categorical cross entropy.

\subsubsection{Compared Methods}
We compare the performance of FedDPC with existing state-of-the-art FL algorithms including FedProx, FedExP, FedGA, FedCM and FedVARP, in which FedProx, FedExp $\&$ FedGA are desiged for handling data heterogeneity and FedCM $\&$ FedVARP are designed for handling both data heterogeneity and partial client participation. The details about these methods is elaborated in section \ref{related_works}.
%%%%%%%%%%%%%%%%%%%%%%%%%%%%%%%%%%%%%%%%%%%%%%%%%%

\begin{figure}[!htb]
  \centering

  \begin{subfigure}{0.85\linewidth}
    \includegraphics[width=\linewidth]{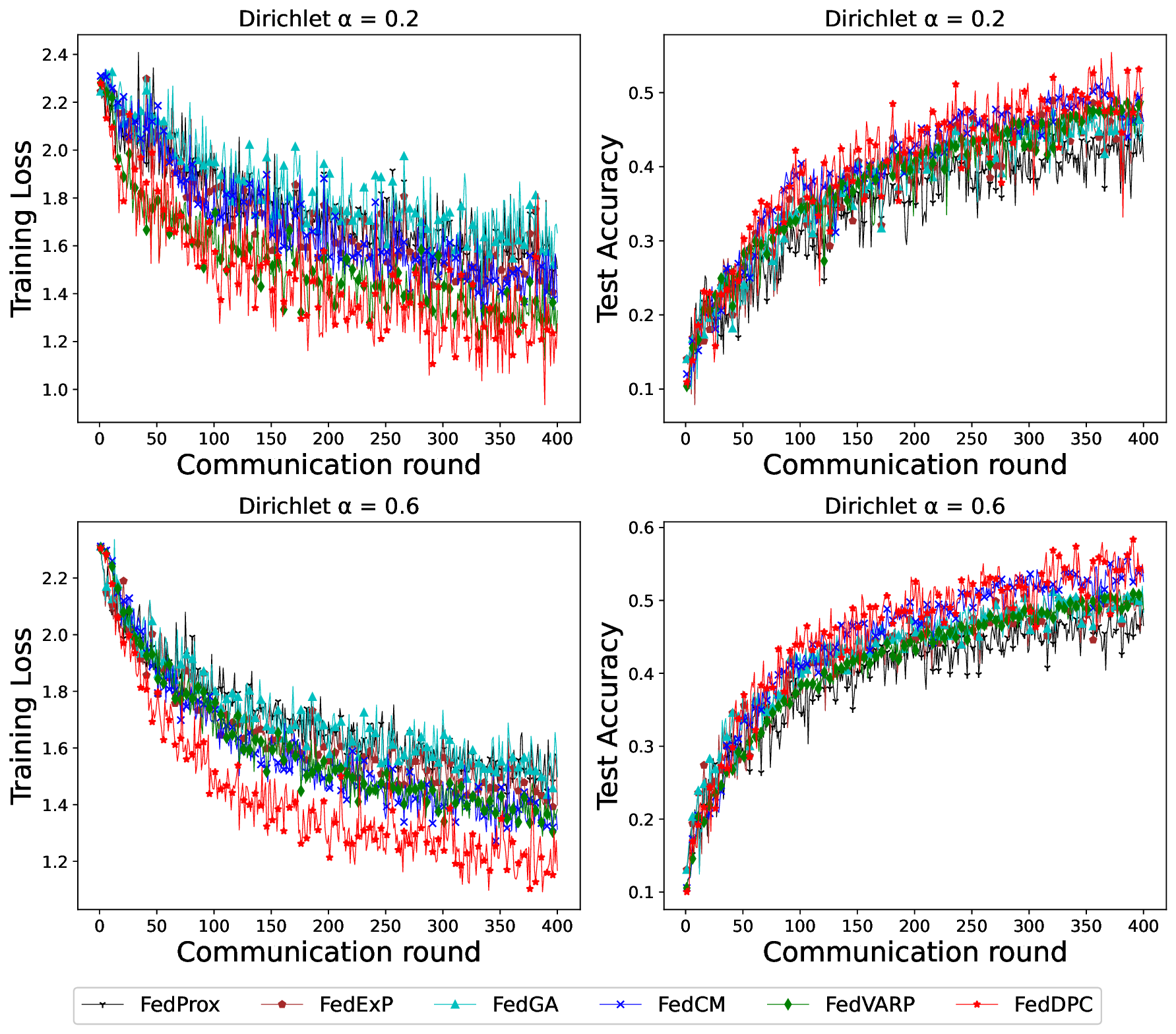} 
  \end{subfigure}
  \caption{Comparison of communication rounds among various methods: Training Loss and Test Accuracy for CIFAR10 image classification using LeNet5 in both the heterogeneous FL settings.}
  \label{fig:1}
\end{figure}

%%%%%%%%%%%%%%%%%%%%%%%%%%%%%%%%%%%%%%%%%%%
\begin{figure}[!htb]
  \centering
  % First row
  \begin{subfigure}{0.85\linewidth}
    \includegraphics[width=\linewidth]{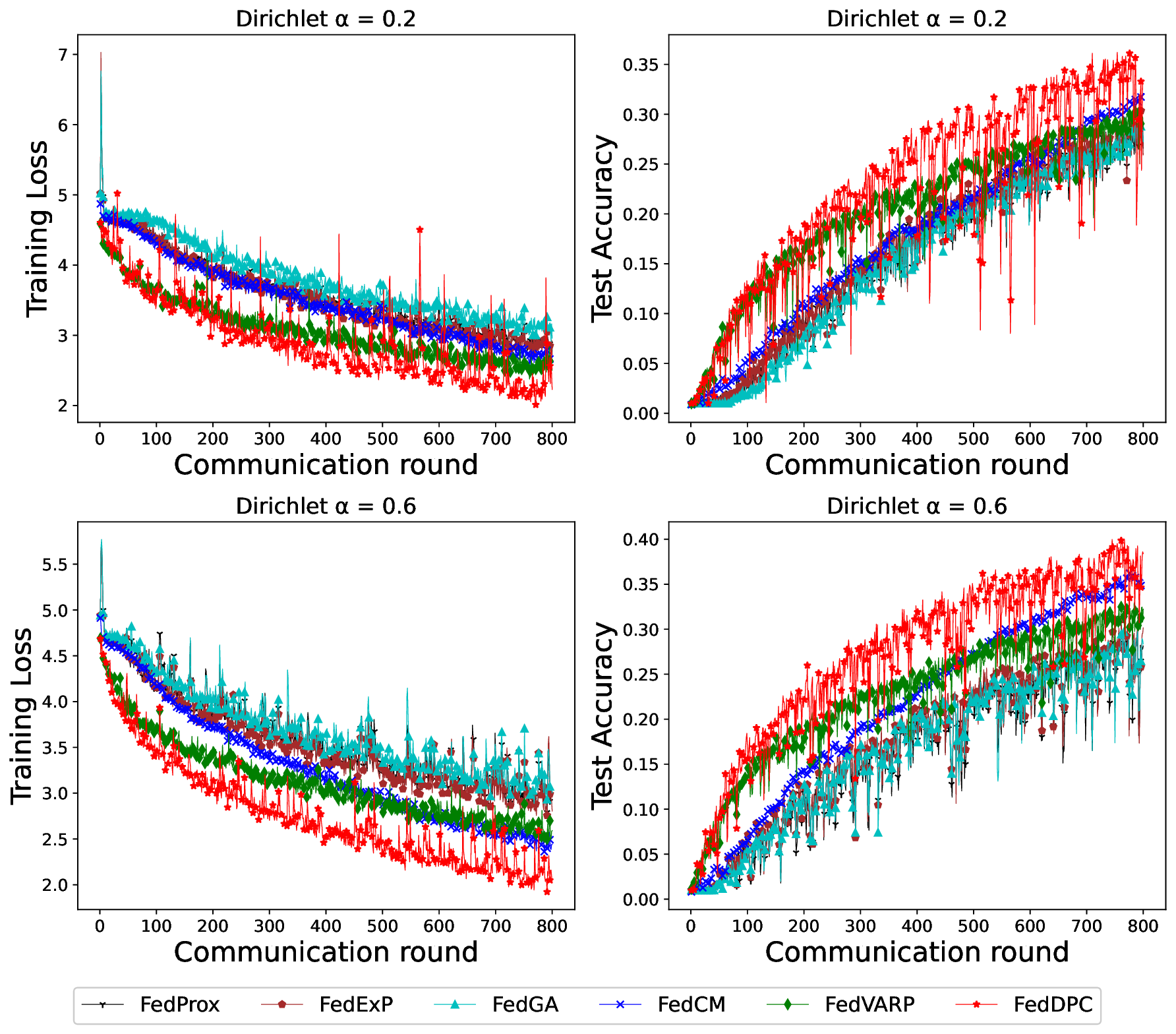} 
  \end{subfigure}
  
  \caption{Comparison of communication rounds among various methods: Training Loss and Test Accuracy for CIFAR100 image classification using Resnet18 in both the heterogeneous FL settings.}
  \label{fig:2}
\end{figure}

\begin{figure}[!htb]
  \centering
  % First row
  \begin{subfigure}{0.85\linewidth}
    \includegraphics[width=\linewidth]{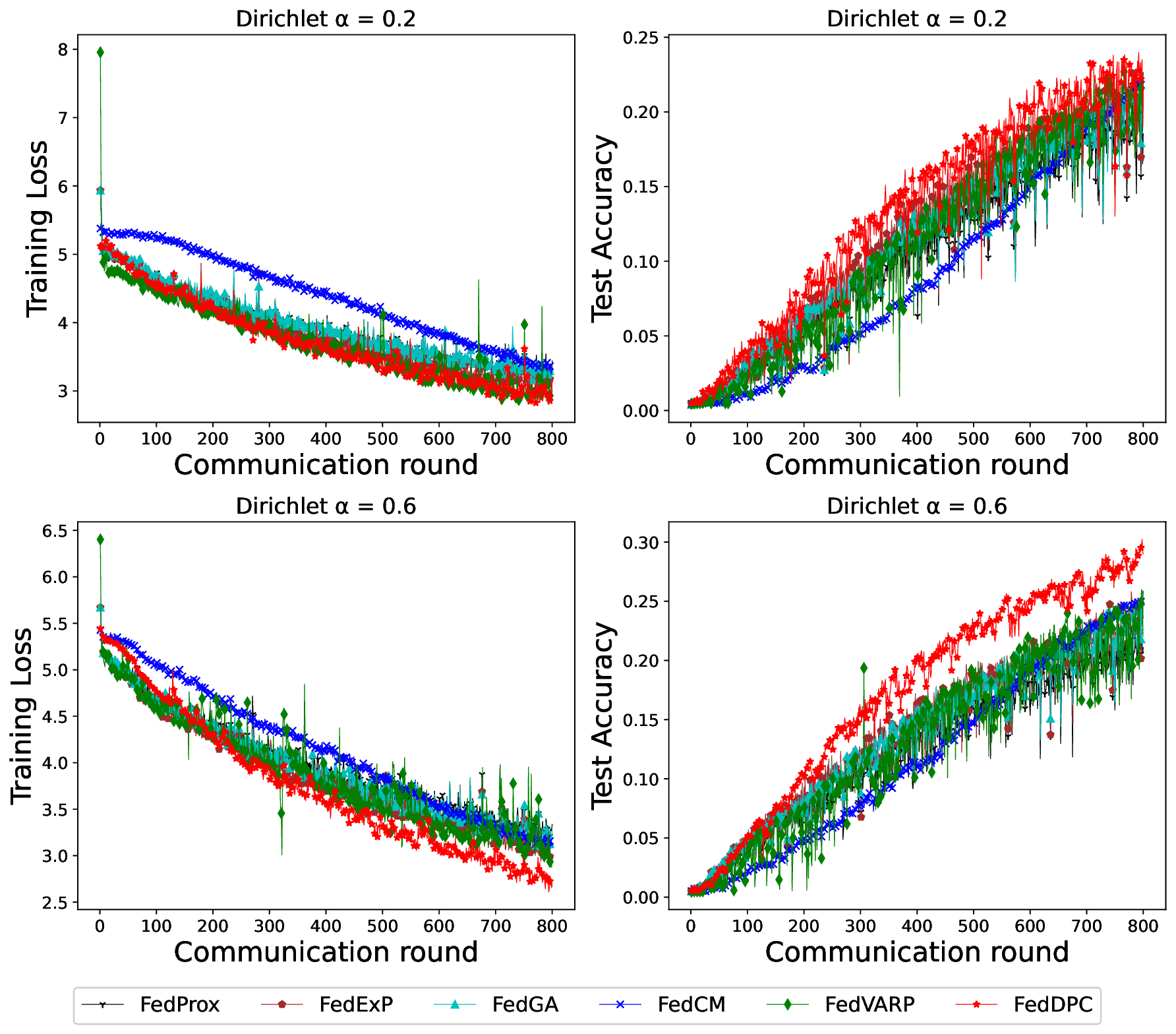} 

  \end{subfigure}
  
  \caption{Comparison of communication rounds among various methods: Training Loss and Test Accuracy for Tiny ImageNet image classification using Resnet18 in both the heterogeneous FL settings.}
  \label{fig:3}
\end{figure}
\subsubsection{Implementation Details}
To find the best performing model for each algorithm, we conduct grid-search tuning and find the best performing model for each algorithm by considering minimum train loss and maximum test accuracy across communication rounds $T$. We use the grid $\{1, 0.1, 0.01, 0.001\}$ for FedProx proximal term $\mu$, FedExP $\epsilon$, FedCM decay parameter $\alpha$ and FedGA displacement step size $\beta$. The grid for learning rate $\eta$ is $\{1, 0.1, 0.01, 0.001, 0.0001\}$ for all the existing and proposed methods. We use $\lambda = 1$ for our proposed method. For local training, we use a batch size of 256 and only one local epoch for all the algorithms. We fix communication rounds $T = 400$ for CIFAR10 and $T = 800$ for CIFAR100 $\&$ Tiny ImageNet. To consider the partial client participation, we randomly sample $10\%$ of total 100 clients in each communication round with the equal probability of participation of each of the clients. We run each algorithm for each dataset with seed value = 0 for reproducibly. While doing experiments on each dataset, we use same initial states and same setting for all the methods to maintain fairness. For implementation, we use Tesla V100 GPU and PyTorch1.12.1+cu102.

%%%%%%%%%%%%%%%%%%%%%%%%%%%%%%%%%%%%%%%%%%%%%
\begin{table*}[h!]
\centering
\small % Use smaller font size
\setlength{\tabcolsep}{6pt} % Reduce column spacing for compactness
\begin{tabular}{*{17}{l}}
\hline
\multicolumn{2}{c}{} & \multicolumn{5}{c}{CIFAR10} & \multicolumn{5}{c}{CIFAR100} & \multicolumn{5}{c}{Tiny ImageNet}   \\
\cmidrule(r){3-7}  \cmidrule(lr){8-12} \cmidrule(lr){13-17}
\multicolumn{2}{c}{} & \multicolumn{2}{c}{$\alpha = $ 0.2} & \multicolumn{2}{c}{$\alpha = $ 0.6} & Time & \multicolumn{2}{c}{$\alpha = $ 0.2} & \multicolumn{2}{c}{$\alpha = $ 0.6} & Time & \multicolumn{2}{c}{$\alpha = $ 0.2} & \multicolumn{2}{c}{$\alpha = $ 0.6} & Time  \\
\cmidrule(r){3-4}  \cmidrule(lr){5-6} \cmidrule(lr){8-9} \cmidrule(lr){10-11} \cmidrule(lr){13-14} \cmidrule(lr){15-16}  
\multicolumn{2}{c}{Method} & Acc & T & Acc & T  & & Acc & T  & Acc & T  & & Acc & T  & Acc & T  &  \\
\hline 
\multicolumn{2}{c}{FedProx}  & 45.20 &317 &48.79 & 378 & 0.50 & 29.03&794&29.90&790&1.05&20.07&794&22.43&786&1.80  \\

\multicolumn{2}{c}{FedExP}  & 49.47 & 363& 51.61 & 399 & 0.45 & 30.35&796&31.07&761&0.92&22.59&781&25.80&779&1.59\\

\multicolumn{2}{c}{FedGA}  & 48.72 & 395& 51.97 & 399 & 0.50& 28.87&797&29.77&790&1.01&21.54&781&24.60&797&1.70 \\ 
\multicolumn{2}{c}{FedCM}  & 52.12 & 369& 56.06 & 378 & 0.49& 32.09&800&36.22&775&0.99&22.00&798&25.21&785&1.72  \\ 
\multicolumn{2}{c}{FedVARP}  & 49.15 & 398& 51.16 & 398 & 0.52& 30.36&790&32.93&784&0.88&22.69&766&25.96&797&1.73  \\ 

\multicolumn{2}{c}{FedDPC}  & \textbf{55.44} & 372& \textbf{57.96} & 364 & 0.51 & \textbf{36.21}&776&\textbf{39.87}&767&0.99&\textbf{23.83}&767&\textbf{30.22}&798&1.65 \\
\hline
\end{tabular}
\caption{Best accuracy (Acc) achieved by different methods in both the heterogeneous settings (Dirichlet $\alpha$ = 0.2 and 0.6) of each dataset within the predefined communication rounds T (T=400 for CIFAR10, T=800 for CIFAR100 and Tiny ImageNet). Here "time" represents the average time in seconds to complete one communication round.}\label{tab:2}
\end{table*}

%%%%%%%%%%%%%%%%%%%%%%%%%%%%%%%%%%%%%%%%%
\subsection{Results}
This section presents our experimental results and compares the performance of the proposed method with existing methods across all datasets, showing results for both heterogeneous settings for each dataset, based on average training loss across clients and test accuracy of the global model. It also includes an ablation study and hyperparameter analysis.

\subsubsection{Comparison with Existing Methods}

The results of our experiments are shown in Figures. $\{\ref{fig:1}, \ref{fig:2}, \ref{fig:3}\}$ and Table \ref{tab:2}. These figures compare the training loss and test accuracy of our proposed method with existing FL algorithms, achieved at different communication rounds, for both heterogeneous settings (Dirichlet $\alpha \in \{0.2, 0.6\}$) on each dataset, with a 10$\%$ client participation rate in each round. From these figures, it may be observed that our proposed algorithm (represented in red) reduces training loss and increases test accuracy more rapidly across different communication rounds compared to existing state-of-the-art FL algorithms, such as FedProx, FedExP, FedGA, FedCM, and FedVARP. Based on these findings, we may conclude that our proposed FedDPC accelerates FL training more efficiently than existing FL algorithms in heterogeneous and partial client settings. This is attributed to the use of same initialization and same settings across all algorithms for each dataset. Table \ref{tab:2} shows that FedDPC achieves the best test accuracy compared to existing methods in both heterogeneous settings for each dataset, with competitive or lower communication rounds than the existing methods. Additionally, this table shows that FedDPC requires a competitive amount of time to complete one communication round compared to existing FL methods, for the FL implementation with 100 total clients. As we conduct experiments on heterogeneous FL settings with partial client participation, we can assert that our proposed algorithm, FedDPC, efficiently mitigates the performance drop of the global model and the slow training of FL, which are caused by data heterogeneity and partial client participation, compared to existing algorithms.
%%%%%%%%%%%%%%%%%%%%%%%%%%%%%%%%%%%%%%%%%%%%%%%%%%%%%%%%%%%
\begin{figure}[!htb]
  \centering
  % First row
  \begin{subfigure}{1.0\linewidth}
    \includegraphics[width=\linewidth]{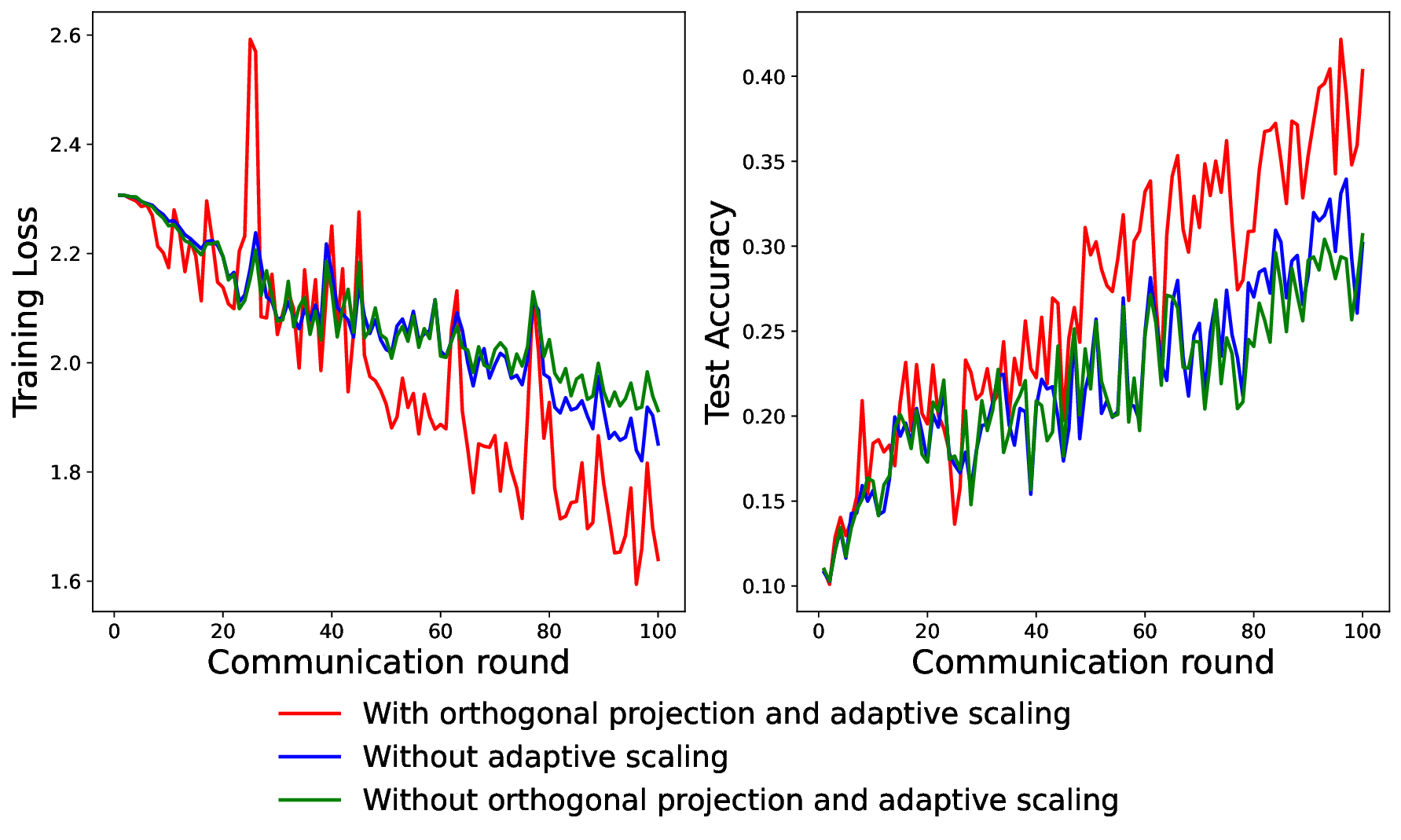} 
  \end{subfigure} 
  \caption{Effects of two key components i.e. orthogonal projection and adaptive scaling, on the performance of FedDPC}
  \label{fig:4}
\end{figure}

\subsubsection{Ablation Study}
In this section, we empirically analyze the effectiveness of the two key components of our designed algorithm, FedDPC: the orthogonal projection-based modification and the adaptive scaling. To assess their individual contributions, we conduct experiments on FedDPC without adaptive scaling and on FedDPC without both the orthogonal projection and adaptive scaling. Since adaptive scaling is applied to the projection-modified local update, it is not feasible to perform experiments on FedDPC without the projection component but with adaptive scaling alone. FedDPC without both components is equivalent to FedAvg with two-sided learning rates. For this study, we perform experiments on federated CIFAR-10 image classification with Dirichlet distribution ($\alpha = 0.2$) using the LeNet5 model, employing the best-performing learning rate (= 0.1) and $\lambda = 1$. The results of these experiments are shown in Figure \ref{fig:4}. From the figure, it may be observed that the projection-based modification alone (represented by the blue line) allows FedDPC to outperform FedAvg with two-sided learning rates in terms of a faster reduction in average training loss across clients and a quicker increase in the global model's test accuracy over communication rounds. This suggests that the projection-based modification in FedDPC improves federated learning (FL) training. Furthermore, the incorporation of adaptive scaling (represented by the red line) accelerates FL training even further by speeding up both the reduction in training loss and the improvement in test accuracy across communication rounds.

\begin{figure}[!htb]
  \centering
  % First row
  \begin{subfigure}{0.8\linewidth}
    \includegraphics[width=\linewidth]{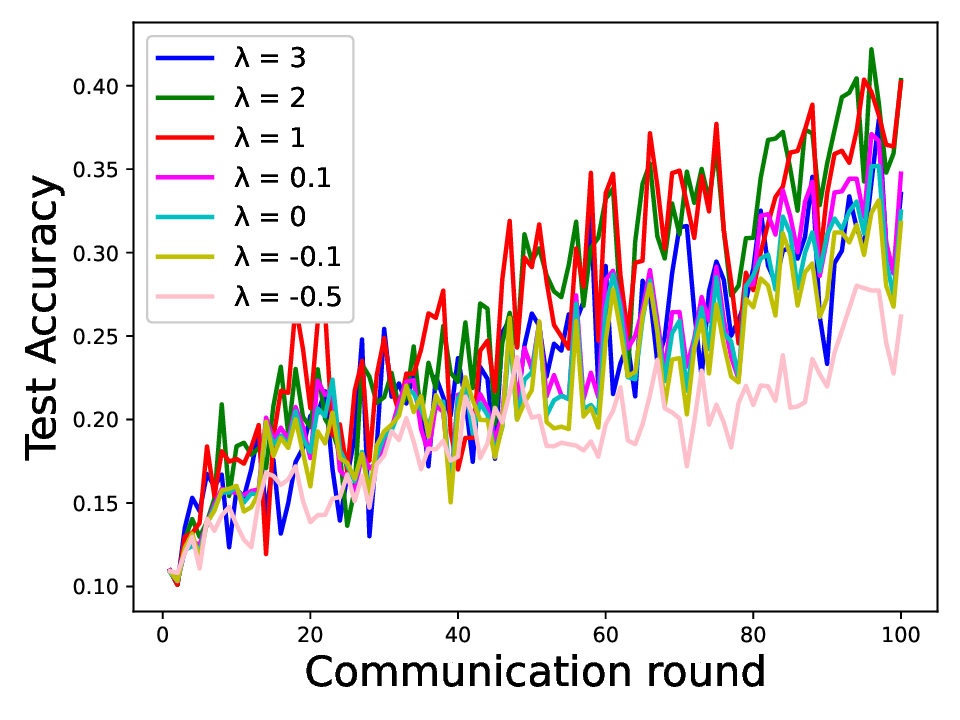} 
  \end{subfigure}
  \caption{Effect of different values of $\lambda$ on the performance of FedDPC}
  \label{fig:5}
\end{figure}

\subsubsection{Effects of Different Values of $\lambda$ on The Performance of FedDPC.}
\label{Sen}
The adaptive scaling of FedDPC is controlled by a hyperparameter $\lambda$. This section analyzes the effectiveness of different values of $\lambda$ on the performance of FedDPC. To perform the same, we conduct further experiments on CIFAR10 dataset (Dirichlet $\alpha = 0.2$) with different values of $\lambda \in \{3, 2, 1, 0.1, 0, -0.1, -0.5\}$ and the best-performing learning rate (= 0.1), as shown in Figure \ref{fig:5}. From this figure, it may be observed that FedDPC performs well in terms of test accuracy across different rounds when $ 0.1<\lambda \leq 2$, and performs very poorly when the value of $\lambda$ is negative. From our experiments on all three datasets, we found that, with $\lambda = 1$, FedDPC yields promising results.

\section{Conclusions and Future Work}
To address the performance degradation and slow training in Federated Learning (FL), this paper proposes FedDPC, designed to mitigate these issues by reducing the variances introduced in local and global updates due to data heterogeneity and partial client participation. This is achieved through orthogonal projection and adaptive scaling techniques. The effectiveness of FedDPC has been demonstrated via extensive experiments on several heterogeneously partitioned datasets with partial client participation, showing its superiority over existing state-of-the-art FL algorithms. In future, we plan to extend this work for vertical federated learning. 
{
    \small
    \bibliographystyle{ieeetr}
    \bibliography{main}
}


\end{document}